# Techniques for Highly Multiobjective Optimisation: Some Nondominated Points are Better than Others

David Corne
School of Mathematics and Computer Science
Heriot-Watt University
Edinburgh, UK
+44(0)131 451 3410
dwcorne@macs.hw.ac.uk

Joshua Knowles
University of Manchester
School of Computer Science
131 Princess Street, Manchester, UK
+44 (0)161 306 4450
j.knowles@manchester.ac.uk

## ABSTRACT
The research area of evolutionary multiobjective optimization (EMO) is reaching better understandings of the properties and capabilities of EMO algorithms, and accumulating much evidence of their worth in practical scenarios. An urgent emerging issue is that the favoured EMO algorithms scale poorly when problems have 'many' (e.g. five or more) objectives. One of the chief reasons for this is believed to be that, in many-objective EMO search, populations are likely to be largely composed of nondominated solutions. In turn, this means that the commonly-used algorithms cannot distinguish between these for selective purposes. However, there are methods that can be used validly to rank points in a nondominated set, and may therefore usefully underpin selection in EMO search. Here we discuss and compare several such methods. Our main finding is that simple variants of the often-overlooked 'Average Ranking' strategy usually outperform other methods tested, covering problems with 5—20 objectives and differing amounts of inter-objective correlation.

## Categories and Subject Descriptors
I.2.8 [**Problem solving, control methods and search**]: *Heuristic methods*

## 1. INTRODUCTION & BACKGROUND
The research area of evolutionary multiobjective optimization (EMO) continues to advance, with the result that we are reaching better understandings of the properties and capabilities of the several algorithms in the field, and accumulating much evidence of their worth in practical and real-world scenarios. An urgent emerging issue is that the favoured algorithms in the field scale poorly when problems have 'many' (e.g. five or more) objectives. This is of concern, since it is not uncommon to encounter real-world problems with 5—20 objectives. Broadly speaking, there seem to be three reasons for this. First, in some algorithms (that perform well when the number of objectives is small), their operation relies on data-structures and subroutines that grow (in size and time respectively) exponentially (or otherwise unreasonably) in the number of objectives. Examples are PAES and PESA [7,19]; in these cases, data-structures and associated algorithms are employed to partition the $k$-objective fitness space into 'hyperboxes', so that a record can be maintained of how many individuals (in the current estimate of the Pareto front, in the population, or both) currently occupy each hyperbox. This forms the basis of certain selection decisions, so that points in less explored regions can be preferred. For very-many objective problems, such techniques become forced to use a small number of divisions per dimension (if we divide each objective into $r$ divisions, there will be $r^k$ hyperboxes), reducing the ability of selection to provide effective discrimination. In the limit, for EMO algorithms (MOEAs) whose selection is based only on such a scheme, their use corresponds to iterated random selection from the (current approximation to the) Pareto front.

The second suspected reason for poor EMOA scaling in terms of $k$ is to do with the preponderance of nondominated solutions as $k$ increases. In simple terms, EMO algorithms such as MOGA [14], NSGA [24], SPEA [27], their sequels, and others, rely on the ability to assign different selective fitnesses to different members of a population, where that population includes a substantial number of dominated points. However, in general terms, many-objective problems are more likely to yield series of populations in which all or most of the points are nondominated. This is because the proportion of a set of random vectors that is nondominated is known to rise quickly with $k$ (e.g. see [1, 2]), suggesting that these algorithms too, in the limit, become equivalent to random selection from the (current approximation to the) Pareto front. What seems needed, therefore, are ways to impose an order of preference over points in a nondominated set.

A third reason for poor scaling in this context may be due to what Hanne has coined 'fitness deterioration' [17]; that is, it will commonly occur, if a size-restricted archive of points is maintained, that points will be lost from the archive at generation $g$, say, that would have dominated points included in the archive at a later (perhaps final) generation. It is intuitive to expect that this phenomenon would be exacerbated in the many-objectives case. We note that there are many complex issues to consider here, but we attempt to make the current work neutral in the sense of the fitness deterioration concept, by employing a simple random archiver. That is, biases in the methods we test arise from the parental selection method, rather than the archiving method.

In many practical scenarios, given domain knowledge and user preferences, we are able to assign at least a partial priority ordering over the objectives. This has long been a successful arm of EMO research [15, 6, 8]. In the many-objective case, such an approach provides ways to distinguish between points on the Pareto front, and may be successful so long as the ordering is valid, but there is always a danger that the sustained favouring of certain objectives may lead the search to miss solutions that may have been preferable to the final result.

In this paper, we take the stance in which one considers all objectives on an equal footing, concerned to find the best possible approximation[1] to the Pareto front. We are therefore interested in methods to obtain a preference ordering over the points in a nondominated set, without recourse to domain-dependent preferences over the objectives. There have in fact been several techniques published that offer such methods, and, in the two earlier cases that we cite, the authors were not particularly (or not at all) concerned with *many*-objective problems. Bentley and Wakefield [3], for example, were interested in the distribution of phenotypes offered by different multi-objective ranking methods, and looked only at 2-objective problems. They compared six such ranking methods: four were of their own design, although one of these was based on VEGA [23], another was non-dominated sorting (first described in [16], and incorporated in [24]), and the other was the equivalent of single-objective fitness (i.e. summing the objective values). In their methods (details appear in section 2.2 for specific methods), any point in a collection $P$ (whether nondominated or not) is considered as a vector of ranks or ratios. E.g. the 2-objective point $p$ may be transformed to (3,7), meaning that it is the $3^{rd}$ best point in $P$ when considering only objective 1, and the $7^{th}$ best when considering only objective 2. The sum of the elements in this vector provides a way to rank the points.

Meanwhile, Drechsler et al [12], also without specific focus on many objective problems, proposed and tested the *favour* relation (although their examples were 6 and 7 objective problems). Like Bentley & Wakefield, they were interested in the ability of their technique to provided a finer grained ordering over multiobjective points than that achieved by the dominance relation alone, and doing so without recourse to any *a priori* preference over the objectives. Essentially, point $s$ is favoured over point $t$ if $s$ is better than $t$ on more objectives than on which $t$ is better than $s$, and if we treat the pairwise favour relations as edges in a graph (and do some necessary processing), we obtain an ordering. Full detail is provided in section 2.2.

More recently, di Pierro [10, 11] has offered the notion of 'efficiency of order $k$', or '$k$-optimality', this time with a distinct motivation to address many objective problems. Essentially, a non-dominated point in $k$ objectives may or may not be dominated if we ignore one or more of the objectives. For example, $a$ = (1, 3, 6), $b$ = (2, 2, 9), $c$ = (5, 4, 5) is a nondominated set of 3-objective points (assuming minimization). However, if we consider objectives 1 and 2 only, point $c$ is dominated by the other two; also, if we consider objectives 1 and 3 only, $b$ is dominated by $a$. Meanwhile, $a$ remains nondominated whatever subset of 2-objectives we choose. We therefore say that $a$ is 2-optimal, while $b$ and $c$ are no better than 3-optimal. Again, further detail and issues with this method are given and discussed in section 2.2.

Another recent method, offered with respect to many-objective problems, is *winning_score* [21]. The idea of the 'Compressed Objective Genetic Algorithm' (COGA) [21] is to treat a many-objective problem as a 2-objective one, where one objective is *winning_score*, which can impose an ordering on nondominated points, and the other is a helper-objective that ensures diversity. COGA is found to be very successful, and we expect in later work to investigate this approach by exploring alternative candidates for the choices of objective to use. Here we were interested in the use of *winning_score* alone, as a comparative rival to other techniques. However, it turns out that *winning_score*, although defined distinctly, is equivalent to one of the methods we use that is taken from [3]; we briefly describe *winning_score* and prove this equivalence in section 2.2.

We note that an alternative direction to take for dealing with the challenges of many objectives is that of *dimension reduction*, currently under research, with different approaches (e.g. [9, 5]). The overall idea in dimension reduction is to find justifiable ways to omit some of the objectives from consideration, choosing such candidates on the basis that their omission will have only minor (or no) effect on the search dynamics and the final solutions reached. In Deb et al's approach [9], this is done by dismissing objectives that are highly correlated with others. Meanwhile, Brockhoff & Zitzler [5] define the *dominance structure* of a set of points, and a metric for evaluating perturbations of that structure, and describes algorithms that can choose a subset of objectives whose dismissal would result in zero or minimal change to the dominance structure. Meanwhile, we note a highly relevant theoretical development from Teytaud [25], in which it is proven that (with certain generally appropriate assumptions) high-$k$ problems with many conflicting objectives become 'too hard' quickly as the number of conflicting objectives rises. That is, lower bounds on the time needed for a MOEA (for example) to find the Pareto set are little better than the time needed for random search in such cases. We find this echoed in our results section, as we consider high-$k$ problems with negative correlations among the objectives. Space limitations preclude a more thorough review (e.g. among other approaches, we have not discussed divide-and-conquer style methods [22], and methods based on probabilistic or fuzzy versions of the dominance concept, e.g. [13]). In general, our overall assessment of the research in this area so far is perhaps not very useful, yet we believe it to be true: whatever method will be best for particular many-objective problems will likely depend crucially on the fitness landscape in question.

Meanwhile, all approaches seem promising, and their investigation will gradually shed light on any limitations, while providing guidelines for the kinds of problem to which they may be best suited. In this paper we begin to contribute to this by investigating the relative abilities of a selection of nondominated-point ranking techniques. We do not cover the full range of techniques available – for example, in the present study we omit consideration of density-based methods that are commonly used in current MOEAs (in which a point's rank, for either parental or environmental selection or both, is increased according to its degree of genotypic or phenotypic isolation from other points). However this enables our conclusions to omit the need for qualification by uniformity issues in the Pareto landscape of the problems under study. Finally, it could be mentioned that one way to distinguish ranking methods concerns whether or not the relative rankings imposed depend on the subset of points in question (e.g. the current population). For example, relative ranks assigned via the *favour* approach depend on the current population (see section 2.2); that is, *a* may be favoured over *b*, or *a* may be deemed equivalent to *b*, depending on what else is in the population. However, if we assign rank according only to the summed objectives, then the relative ranking of two distinct points will be the same in different populations. Among the methods we examine are examples of each.

---

[1] There is no fully accepted definition of 'best possible approximation' to the Pareto front; the difficulty of the issue is reflected in the diversity of associated metrics available, and the burgeoning literature on the topic of metrics itself. As ever, what we mean in this case by 'best possible approximation' is subsumed in our choice of metric.

The remainder is structured as follows. Section 2 contains some preliminaries, introducing our notation, and describing a collection of published methods for ranking non-dominated points; we include a simple proof of equivalence between two of these methods, and finally describe our test problems. In section 3 we describe preliminary investigations into the rank distributions achieved by favour, and *k*-optimailty. In section 4 we describe a large collection of experiments that compare the methods described in section 2, together with some baselines. Section 5 summarises and concludes.

## 2. DEFINITIONS & NOTES
### 2.1 Preliminaries

Given a search space of structures $S = \{s_1, s_2, ..., s_{|S|}\}$, and $k$ *objective* functions $f_1, f_2, ..., f_k$, each with domain $S$ and range $\Re$, and then given the need to find structures in $S$ that 'simultaneously' minimize each of our $k$ functions, we are faced with a *multi-objective problem* (MOP). In practice some or all of our $k$ functions may need to be maximized, but it is trivial to convert such cases and treat them all as minimization problems without loss of generality.

Given any subset $P \subseteq S$ (such as the population, at some generation $g$, in a population-based algorithm attempting to solve this MOP), it will contain, for each objective $j$, one or more solutions that are best (i.e. minimal for that objective) when compared with the other individuals in $P$. Occasionally, a single individual may be best for all objectives, but commonly this is not the case. The key relationship between individuals in this context is *dominance*. We say that $s_i$ dominates $s_j$ in the case that $s_i$ is better than $s_j$ on at least one objective, while $s_j$ is not better than $s_i$ on *any* objectives. Formally:

$s_i$ dominates $s_j \Leftrightarrow$
$\forall o \in [1,...,k], f_o(s_i) \leq f_o(s_j)$ AND $\exists q \in [1,...k], f_q(s_i) < f_q(s_j)$

A similar relationship, *coverage*, also comes in useful. We say that $s_i$ covers $s_j$ in the case that $s_i$ is not worse than $s_j$ on any objective; i,e., either $s_i$ dominates $s_j$, or the two are equal (in objective space). Formally:

$s_i$ covers $s_j \Leftrightarrow \forall o \in [1,...,k], f_o(s_i) \leq f_o(s_j)$

Our final preliminary is to define the *nondominated set* with respect to our set $P$. This is simply those in $P$ that are not dominated by any others in $P$.

Now it is time to consider how we might assign selective fitnesses to points in $P$. In all of the following, although we have defined $P$ as a set of unspecified structures from a search space $S$ (e.g. perhaps they are graphs, neural networks, and so on…), we will treat them as if they are characterized by the vector of $k$ objective values that they are mapped onto by our objective functions.

### 2.2 Ranking Nondominated Points

Consider only the nondominated set from $P$, i.e. $N = \text{nd}(P)$. In practice, we have no clear *a priori* way to decide which may be 'best' among any two points in $N$; if such a distinction needs to be made, the decision-maker will choose based on domain knowledge and pragmatics. However, there are some approaches that seem to have potential merit for *a priori* distinctions. In historical order (in our current understanding) these include the following:

#### 2.2.1 OBJECTIVE RANKING AND RATIOS

In [3], the following techniques are defined, all capable of inducing a preference ordering over a set of nondominated points: *weighted average ranking* (WAR), *weighted maximum ranking* (WMR – a basic extension to VEGA [23]), *sum of weighted ratios* (SWR), and *sum of weighted global ratios* (SWGR). Bentley and Wakefield were partly interested in the use of these techniques in cases where *a priori* preferences existed for the objectives, which could be expressed as weights, but here set all weights to 1, and accordingly omit this aspect from the names and abbreviations.

Some further notation will be helpful for this and some later sections. A much simpler elaboration is possible, but the treatment here will support a simple proof later of the equivalence between AR and *winning_score* [21]. To that end, consider the three-dimensional matrix $\mathbf{A}$, such that:

$$a_{ijk} \in \mathbf{A} = \begin{cases} 1, & \text{when } f_k(s_i) < f_k(s_j) \\ 0, & \text{when } f_k(s_i) = f_k(s_j) \\ -1, & \text{when } f_k(s_i) > f_k(s_j) \end{cases}$$

In words, $a_{ijk}$ records 1, 0, or −1, depending on whether $s_i$ is better, equal to, or worse than $s_j$ on objective $k$. The AR method calculates a score for each point $s_i$ by summing the ranks of $s_i$ for each objective. E.g. if there are 3 objectives, and $s_i$ is 2$^{nd}$ best on two of these and 5$^{th}$ best on the other, its AR score, $AR(s_i)$, will be 2+2+5 = 9. It is easy to see that, for $s_i \in P$:

$$AR(s_i) = \sum_k \sum_{j \neq i} (|P|+1) - a_{ijk}$$

The inner sum calculates a score for $s_i$ for a given objective, and this will be 1 if $s_i$ is the best on that objective, and generally $z+1$ if $z$ members of $P$ are better on that objective.

Meanwhile, the SR method simply replaces the rank for a given objective (e.g. "3$^{rd}$ best") with the normalized objective value. That is, given a set of *k*-objective points $P$, assuming minimization in each objective, we have:

$$SR(s_i) = \sum_k \text{nratio}(s_i, k)$$

where:

$$\text{nratio}(s_i, k) = \frac{f_k(s_i) - \min_{s \in P}\{f_k(s)\}}{\max_{s \in P}\{f_k(s)\} - \min_{s \in P}\{f_k(s)\}}$$

We considered AR and SR the more promising of the methods from [3] to test, and do not from hereon discuss or consider the others, except to provide a brief definition for the interested reader. First, SGR is the same as SR, but in which the normalizations are done with respect to all points found so far during the search, rather than those in the current population. Second, MR takes the *best*, rather than the average (or sum) of ranks for each objective.

*2.2.2 The Favour Relation*

Drechsler et al [12] proposed the following idea, as a way to be able to provide distinctions between points in a nondominated set. Let $N$ be a nondominated set of $k$-objective points. We will say that we favour $s_i$ over $s_j$ (i.e. $s_i$ favour $s_j$) in the following situation:

$s_i$ favour $s_j \Leftrightarrow$
$A = \{o : f_o(s_i) < f_o(s_j)\}, B = \{o : f_o(s_i) > f_o(s_j)\}, (|A| - |B|) > 0$

In words, we favour $s_i$ over $s_j$ if $s_i$ is better than $s_j$ on more objectives than in which $s_j$ is better than $s_i$. In making use of this relation, the idea is to induce an ordering on the points in $N$. However, favour is not transitive. For example, consider these six-objective points: $p = (0, 1, 2, 3, 4, 5)$, $q = (1, 2, 2, 3, 4, 0)$, $r = (3, 4, 2, 3, 0, 0)$, $s = (4, 5, 2, 0, 0, 0)$. We have $p$ favour $q$, $q$ favour $r$, $r$ favour $s$, and $s$ favour $p$, and so we cannot determine an ordering over this set. Drechsler et al's approach is to consider the directed graph induced by the favour relation, and collapse such cycles into so-called 'strongly connected components' (SCCs). The overall method, for ranking points in $N$ is:

1. Determine the *favour graph*, composed of directed edges $p \rightarrow q$ for all pairs of points where $p$ favour $q$.
2. Collapse all cycles in the favour graph into individual SCCs (i.e. treat each SCC as a single node).
3. Determine a partial ordering on the basis of the resulting DAG.

As pointed out in [12], determination of the SCCs in the favour graph can be done in linear time. But there are concerns with regard to this relation. Since it is not transitive, it is possible for the entire favour graph to 'collapse' and yield no grounds at all for selective discrimination between the points. Meanwhile, in many objective problems, intuition suggests that cycles in the graph could be numerous, and hence, if not collapsing entirely, it may be common to find that points become assigned to relatively few ranks, again with an impoverished level of selective discrimination. We investigate such issues in our first set of experiments.

*2.2.3 K-Optimality*

The most recent novel approach to finding an ordering over nondominated points is di Pierro et al' '$k$-optimality'. At the risk of confusion, we will continue to retain $k$ to indicate the total number of objectives, and define this as follows.

Given the point $s$ in a nondominated set of $k$-objective points $N$, $s$ is efficient of order $z$, where $1 \leq z \leq k$, iff $s$ is nondominated in every $z$-objective subset of the $k$ objectives (a simple example was given in Section 1). Certain salient properties are as follows [10]. Every point in $N$ is efficient of order $k$, by definition, and: if $s$ is efficient of order $z$, where $z<k$, then $s$ is also efficient of order $z+1$ Finally, we say that $s$ is '$z$-optimal' when $z$ is the lowest value for which we can say $s$ is efficient of order $z$. To use this as an approach to rank nondominated points, we simply associate each point $s$ with rank $z$, such that $s$ is $z$-optimal.

Di Pierro et al [10, 11] have found good results using this method, however it is not clear, without experimentation, how useful a ranking will typically be provided by this method. It is easily noted that there can be at most $k-1$ distinct ranks in any set of nondominated points (notice that for a point to be 1-optimal, the set must be a singleton), although di Pierro et al also describe [10] a finer-grained version of the method that we do not explore here. It should also be noted that we have not yet found a efficient way to determine $z$-optimality, and so the method becomes unusable beyond around 20 objectives.

*2.2.4 Equivalence of Winning Score and AR*

Finally, we briefly note the *winning_score* (WS) technique [21], presented within the 'Compressed Objective Genetic Algorithm' (COGA). Results using WS [21] led us to consider it for this study, and we here present the definition. Given a set of nondominated points $N$, the WS rank of a point $s$ is the sum of its 'margins' over all other points. For example, if $s$ is better than $t$ in 3 objectives, but worse than $t$ in 1 objective, its margin over $t$ is 2 (and $t$'s margin over $s$ is $-2$). Using notation from section 2.2.1, and rmaintaining the convention of lower ranks indicating better points (hence the minus sign below), we can formalise this as:

$$WS(s_i) = -\sum_{j \neq i} \sum_k a_{ijk}$$

However, note that

$$AR(s_i) = \sum_k \sum_{j \neq i} (|P|+1) - a_{ijk} = C \sum_k \sum_{j \neq i} a_{ijk}$$

for a constant $C = (|P|+1) \cdot k \cdot (|P|-1)$, while by simply switching the order of summations we can confirm that:

$$AR(s_i) = C \cdot WS(s_i)$$

and so the orderings induced by these two methods are the same.

Finally, we note that methods very similar to those discussed here have been long considered in the field of multicriteria decision-making (MCDM), e.g. see [18]. In that area, however, the emphasis is on finding compromise points, rather than finding approximations to the full Pareto set.

## 2.3 Metrics and Test Problems

*2.3.1 Relative Entropy*

Given two methods, *A* and *B* respectively, for assigning ranks to a set of 100 nondominated points, we can compare *A* and *B* in terms of the distribution of ranks induced. For example, suppose method *A* assigns rank 1 to one of the points, and rank 2 to the remaining 99 points; meanwhile suppose that method *B* assigns rank 1 to 30 of the points, rank 2 to another 30, and rank 3 to the remaining 40. *B* would seem to have provided a better result, since it has yielded a richer ordering, providing more for selection to 'bite' on. For reasons briefly discussed in sections 2.2.2 and 2.2.3, we are interested in assessing the behaviours of both *favour* and *k*-optimality in this respect. To that end, we will assess rank distributions in terms of their *relative entropy*.

Consider a set of points *N*, each of which has (w.l.o.g.) a positive integer rank (there are at most |*N*| ranks, and at least 1). The distribution of ranks *D* is as follows: *D*(*r*) gives the number of points from *N* with rank *r*. Its relative entropy is:

$$\text{re}(D) = \frac{\sum_r \frac{D(r)}{|N|} \log(\frac{D(r)}{|N|})}{\log(1/|N|)}$$

This becomes close to (reaches) 1 as we approach (reach) the 'ideal' situation in which the points are totally ordered over |*N*| distinct ranks. It is zero in the case where the distribution simply gives every point the same rank.

### 2.3.2 MOTSPs and MOSMJSPs

We use two kinds of test problems: multiobjective traveling salesperson problems (MOTSPs – hereafter, simply TSPs) and multiobjective single-machine job-shop problems (JSPs). Our *k*-objective TSPs simply comprise *k* distance matrices, one for each objective. We generate one as follows. We first generate the TSP for objective 1 by assigning each distinct pair of cities with a uniform random number between 0 and 1. Then the TSP for objective *i*+1 is generated as follows for each entry in the matrix:

$$\text{distance}_{i+1}(a,b) = \text{TSPcp} \cdot \text{distance}_i(a,b) + (1 - \text{TSPcp}) \cdot \text{rand}()$$

where TSPpc, $-1 < \text{TSPcp} < 1$, is a simple TSP 'correlation parameter'; when less than 0, 0, or greater than 0 respectively, it introduces negative, zero, or positive interobjective correlations. In every case reported here, we use 30-city TSPs.

Meanwhile, in a *k*-objective *n*-job such problem, each job has a processing time, and a due date, and each has to be processed on the same machine. The chromosome therefore (just as with the TSP) specifies a permutation of the *n* jobs, which represents their order of processing. Consider the simple case of three jobs, *A*, *B* and *C*, with processing times respectively 20, 30, 40, and due dates 50, 20, 60. The chromosome *CAB* represents the case where *C* is processed first, hence finishing at time 40, earlier than its due date. Because it finishes no later than its due date, *C*'s *lateness* is 0. Job *A* then begins at time 40 (the earliest time it can now start) and finishes at time 60; its due date is 50, so *A* has a lateness of 10. Finally, *B* starts at 60 and finishes at 90, with lateness 70.

In our test SMJSPs, the *n* jobs are each assigned uniformly at random to one of *k* customers. Objective *i*, $1 \leq i \leq k$, is the sum of the latenesses of the jobs assigned to customer *i*. In each case, the job processing times are assigned uniformly at random between 50 and 200, and the due dates are assigned uniformly at random between 50 and 150×JSPcp, where JSPcp is our simple correlation parameter for these problems. Higher values of JSPcp provide more chance of solutions existing where many or all problems have zero lateness, hence with potentially small or vanishing Pareto fronts; lower TSPcp values induce greater conflict, and hence negative inter-objective correlations.

### 2.3.3 MOO Comparison Metric

Finally, we note that we use the standard *cover metric* to compare the performance of two MOEAs [27]. Where *A* and *B* are the archived solutions arising from two algorithm runs, *Cov*(*A,B*) indicates the percentage of set *B* that is covered (see section 2.1) by points in *A*, and *Cov*(*B,A*) is appropriately defined *vice versa*. In a two-algorithm comparison, we simply take *Cov*(*A,B*) > *Cov*(*B,A*) to indicate a 'win' for (the algorithm that produced) *A*, and count a series of such tests as statistically significant if one or other algorithm was the winner suitably often. This is not ideal or authoritative (there is a growing literature on performance metrics for MOEAs, e.g. see [10, 26, 28]); however, early investigations showed that, in the case of the experiments done here, the cover metric provided sufficient distinctions in performance in this case.

## 3. RANK DISTRIBUTIONS

In the context of many-objective population-based search, we would like the following to be true: the better a point is ranked, the more likely it is that selection of this point will lead to (via crossover and/or mutation operations) improvements in our current approximation to the Pareto front. For the sake of discussion, we will call this correlation between rank and 'Effective Fertility' the *EF-quality* of the ranking. If a ranking method provides a low relative-entropy distribution, then it seems not to provide much discrimination between points in this sense, so we would suppose that a higher relative entropy would lead to a better performing MOEA. But a high-relative entropy distribution may well have low EF-quality. Relative entropy is therefore a rough guide to the effectiveness of a ranking method, becoming increasingly unreliable with higher values.

Nevertheless, we felt it worthwhile to preliminary study on the distributions induced by favour and *k*-optimality. In the case of favour, it seems likely *a priori* that many-objective populations will have rather low relative entropy, and we were interested in the degree to which zero-scoring distributions occurred. In the case of *k*-optimality, and given that a maximum of *k*–1 ranks are possible, we were interested in the degree to which its distributions were better (in relative entropy terms) than those of favour. As for AR and SR, we simply note that these tend to produce distributions with maximal relative entropy (as intuition would indicate) – i.e., usually, each point in the population will have a different rank – inducing a total ordering. So, AR and SR are not further investigated in this section.

We performed the following experiments: for each number of objectives *k* in {5, 6, 7, 8, 9, 10, 12, 14, 16, 18, 20}, we repeated the following 1,000 times:

1. Generate and evaluate a random population of 50 *k*-objective MOTSP individuals.

2. Calculate the relative entropy of the *favour* method on this population.

3. Calculate the relative entropy of the *k*-optimality method on this population.

Figure 1 summarises the results.

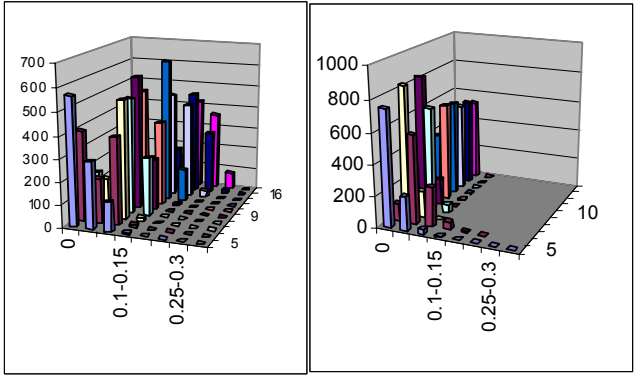

**Figure 1. Distributions of relative entropy values for 1000 random populations ranked by *k*-optimality (right) and favour (left). Leftmost bar in each distribution indicates the number of zero relative-entropy results; subsequent bars indicate frequency of results with r.e. in (0,0.05], (0.05,0.01], and so on until (0.95,1.0]. Into the page, *k* goes from 5 to 18.**

It is striking that favour induces very low r.e. distributions, especially as *k* increases. In contrast, *k*-optimality provides reliably above-zero r.e. values, improving with *k*. Expectations that we can draw from this are as follows. Favour seems unlikely to provide an effective way to discriminate between nondominated points, simply because it rarely provides a rank distribution that discriminates between the points at all. However, we might still be surprised if it turns out that the EF-quality of any favour-induced ranking is high. Meanwhile, *k*-optimality seems to induce a modest but steady rank distribution, which we would expect to lead to better performance than favour, especially since published results seem to testify to its EF-quality [10, 11].

## 4. COMPARATIVE EXPERIMENTS

We performed experiments to evaluate the relative quality of the four nondominated-point ranking methods described. We also tested three baseline methods; all seven are listed below:

- ARF: Bentley & Wakefield's (Weighted) Average Ranking [3], confined to Pareto front points only (i.e. ranks are calculated according to the PF of the current population only, and only PF points are selected).
- SRF: Bentley & Wakefield's Summed (Weighted) Ratio [3], again confined to PF only.
- FR: The favour relation [12]; again confined to PF.
- KO: *k*-optimality [10] (only defined for PF points)
- RF: random selection, but confined to PF.
- SO: selecting by single-objective (sum of objectives) fitness, where each objective is normalized with respect to the current population (hence eliminating biases induced by objective ranges). In fact this is equivalent to SR (SRF not confined to PF points).
- RR: random selection.

In all experiments, we used the following MOEA, designed for simplicity. The archive size was 100, and was maintained in a simple way, to avoid confusing the results with potential interactions between specific archiving strategies and the specific methods. As customary, a point always entered the archive if not already covered the archive, and if the archive was not already full. Naturally, if a new archive point dominated any existing ones, these latter were removed. When a point was not covered by the archive, but the archive *was* full, the point always entered the archive, and a random point was removed. The EA itself was as follows, with parameter settings given within the pseudocode

1. Initialise: generate a set 20 of popsize random individuals, and evaluate each. Find their Pareto front $F$, and set archive $A = F$
2. Repeat 500 times:
   a. Determine ranks for each point in the population $P$ according to METHOD
   b. Initialise intermediate generation $I$ with a random point from $F$.
   c. Repeat popsize-1 times:
      i. Select a parent from METHOD.SET using tournament selection (tournament size 5) over rank.
      ii. Produce mutant $m$ via adjacent-swap mutation. Evaluate and archive $m$, then add it into $I$.
   d. Replace $P$ with $I$.

### 4.1 Experiments and Results

The MOEA of section 4.2 was run 20 times with different random seeds for all 700 combinations of the following settings:

METHOD $\in$ {AR, SR, FR, KO, RF, NSO, RR}

$k \in$ {5, 10, 15, 20}

TSPcp $\in$ {−0.4, −0.2, 0, 0.2, 0.4}

JSPcp $\in$ {10, 20, 30, 40, 50}

The MOTSPs each had 30 cities, and the SMJSPs each had 30 jobs, each of which was randomly assigned to one of *k* customers.

An individual trial involved running all 7 methods for a specific combination of *k*, TSPcp, and JSPcp. After 20 such independent trials for each combination, we computed the cover metrics for each of the resulting 28 distinct pairs of archives. Hence, for a given set of 20 trials, and for each pair of methods $A$ and $B$, we have 20 paired values for Cov($A,B$) and Cov($B,A$). With the hypothesis of no difference between $A$ and $B$, we note that 17 or more occurrences of Cov($A,B$) > Cov($B,A$), or vice versa, occurs with probability <0.0013. Allowing for the simple Bonferroni correction [4] this corresponds to a respectable *p* value of 0.036 per individual comparison. We will take 17 'wins' to be our threshold for significance in the following. We present a summary of the TSP results in Table 1, whose interpretation is as follows.

Each row summarises the result for comparing a pair of methods. Consider the ARF vs RF row, column *k*=20. The five characters isummarise the results for each of the five TSPcp values in the order given at the start of this section. Their meanings are: A – the first method (in this case ARF) was better, with statistical significance; 0 – neither method won enough Cov($A,B$) *vs* Cov($B,A$) comparisons to achieve significance; **B** – the second method (in this case RF) was better, with statistical significance. So, in this case, RF was superior to ARF on the highly negatively correlated 20-objective TSP problem, but ARF was superior to RF for all other values of TSPcp in this 20-objective case.

**Table 1. Summary of TSP results. See text for explanation.**

| Comparison | k=5 | K=10 | k=15 | k=20 |
|---|---|---|---|---|
| ARF vs SR | AAAAA | AAAAA | AAAAA | BAAAA |
| ARF vs FR | AAAAA | AAAAA | AAAAA | AAAAA |
| ARF vs KO | AAAAA | AAAAA | AAAAA | AAAAA |
| ARF vs RF | AAAAA | AAAAA | AAAAA | BAAAA |
| ARF vs SO | AAAAA | AAAAA | AAAAA | BAAAA |
| ARF vs RR | AAAAA | AAAAA | AAAAA | BAAAA |
| SR vs FR | BBBBB | BBBBB | 00BBB | AA00B |
| SR vs KO | BBBBB | BBBBB | 00BBB | AA0BB |
| SR vs RF | BBBBB | BBBBB | 00BBB | 000BB |
| SR vs SO | 0BBBB | 0000B | 00000 | 00000 |
| SR vs RR | BBBBB | BBBBB | 000BB | 000BB |
| FR vs KO | 00BBB | BBBBB | 00BBB | 00BBB |
| FR vs RF | 000BB | BBBBB | 00BBB | BBBBB |
| FR vs SO | AAAAA | AAAA0 | 000AA | BB0AA |
| FR vs RR | AAAAA | BBBBB | 0000B | BBBBB |
| KO vs RF | 000AA | 0AAAA | 0000A | 00000 |
| KO vs SO | AAAAA | AAAAA | 00AAA | BB0AA |
| KO vs RR | AAAAA | 0AAAA | 0000A | BB000 |
| RF vs SO | AAAAA | AAAAA | 00AAA | 000AA |
| RF vs RR | AAAAA | 0000A | 00000 | 00000 |
| SO vs RR | BBBBB | BBBBB | 000BB | 000BB |

Table 1 achieves a concise summary, but is not helpful for an 'at-a-glance' appreciation of the findings. To help, Table 2 provides a rank-ordering (best to worst). In each case, if algorithm A is to the left of algorithm B in the ordering, it was not bested with statistical significance by any on its right, given the experimental design context of the row in question.

**Table 2. Rank orderings of the methods.**

| Expts | Rank-Ordeering |
|---|---|
| TSP 5 / -40 | ARF, FR, KO, RF, RR, SO, SR |
| TSP 5 / -20 | ARF, FR, KO, RF, RR, SO, SR |
| TSP 5 / 0 | ARF, KO, FR, RF, RR, SO, SR |
| TSP 5 / 20 | ARF, KO, FR, RF, RR, SO, SR |
| TSP 5 / 40 | ARF, KO, FR, RF, RR, SO, SR |
| TSP 10 /-40 | ARF, KO, RF, RR, FR, SR, SO |
| TSP 10 /-20 | ARF, KO, RF, RR, FR, SR, SO |
| TSP 10 / 0 | ARF, KO, RF, RR, FR, SR, SO |
| TSP 10 / 20 | ARF, KO, RF, RR, FR, SO, SR |
| TSP 10 / 40 | ARF, KO, RF, RR, FR, SR, SO |
| TSP 15 / -40 | ARF, {all others equally rated} |
| TSP 15 / -20 | ARF, {all others equally rated} |
| TSP 15 / 0 | ARF, KO=RF, FR, SR = SO= RR |
| TSP 15 / 20 | ARF, KO=RF, RR=FR, SR=SO |
| TSP 15 / 40 | ARF, KO, RF=RR, FR, SR=SO |
| TSP 20 / -40 | SO=RR=SR, RF, ARF, KO=FR |
| TSP 20 / -20 | ARF, NSO=RR=SR, RF, KO=FR |
| TSP 20 / 0 | ARF, KO=RF=RR, FR=SO=SR |
| TSP 20 / 20 | ARF, KO= RF=RR, FR, SO=SR |
| TSP 20 / 40 | ARF, KO=RF=RR, FR, SO=SR |

Lack of space prevents showing these tables for the SMJSPs, so we simply describe the main differences here: the findings were very similar, with differences that we can relate to the respective calibrations of the correlation parameters; generally, JSPs were more negatively correlated, which meant less clear discrimination between methods as we increased $k$, and reduced JSPcp. Hence, for example, the $k$=15 column for the JSPs resembled the $k$=20 column in Table 1, while both the $k$=20 and k=15 columns for the JSP table contained more 0s than their TSP counterpart. As for table 2, the JSP results show a very similar story, with most differences in the $k$=20 problems, in which results such as in the TSP 20/-40 column in Table 2 occurred in each case.

The striking result is that ARF outperforms the other algorithms in most cases, defeated only when there are both very many objectives and high conflict ($k$=20 and TSPcp = -40; $k$=15,20 and JSPcp = 10, 20). Notably, in precisely these cases, no method was superior to random search (RR), reflecting the predictions in [25]. Though favour performed fairly well when $k$ was 5 or 10, the other clear good performer was $k$-optimality, in most cases second only to ARF. It is interesting that certain algorithms performed consistently worse than RR. These were SO, SR, and (for 10 or more objectives) FR. An explanation is that the biases induced by the rankings in these methods work against well-distributed progress in the Pareto front as the size of the latter increases. E.g. SO will naturally favour compromise points central to the front, concentrating on an area of the space that proportionally vanishes as $k$ and negative-correlation increase. The relatively good performance of RF adds support to such an explanation.

## 5. CONCLUDING DISCUSSION

Research into *many*-objective optimization problems is growing in necessity. One route towards dealing effectively with such problems is to find ways to induce a preference ordering over nondominated points, such that this ordering is effective for selection purposes. We have compared several methods for this task, and found that the average ranking (ARF) method is highly effective in comparison with the other methods, as long as (loosely speaking) the objectives do not exhibit any significant inter-correlation. The next-best method was $k$-optimality [10], while favour and random selection from the Pareto front both tended to do well, respectively on small numbers and high numbers of objectives. Tentative beginnings of explanations are presented above in terms of the biases introduced by different ranking methods, but we believe that further investigations into our vague notion of 'effective fertility', and how it correlates with different ranking methods, may shed more light. For example, we suspect that $k$-optimality's rankings correlate well with 'effective fertility', since it tends to perform very well *despite* inducing distributions with only modest entropy. We also suspect that AR's ranking approximates that induced by $k$-optimality, but providing even finer grained distinctions.

Our main finding is that ARF, based on the somewhat overlooked AR method [3] seems a very strong candidate for many-objective search. It seems to outperform two recent methods that have been proposed specifically for many-objective problems. Usefully, it is also computationally simple and efficient.. Perhaps because it was not published in mainstream (MO)EA literature, and also perhaps because the comparisons were not based on quality measures of the Pareto set, and also perhaps because it tends to be associated with objective-priority approaches, Bentley & Wakefield's WAR has been little used, omitted for example in the several large or moderate scale MOEA comparison papers. From our results, we recommend, of course, that ARF be further tested and used in many-objective scenarios; but we also suggest both ARF and AR be revisited as candidates for 'standard' (2—5 objective) problems; preliminary studies indicate that it performs comparatively well in such cases too.

Finally, we point out various limitations. Although two test problems are better than one, we cannot do better than generalize very tentatively from the MOTSP and SMJSP problems. Similarly, our use of a single comparative performance metric is less than *de rigeur*. More interestingly, we note that alternative versions of favour [12] can be constructed in which $s$ is said to favour $t$ in the case that $s$ is better than $t$ in more than $n$ (>0) objectives. By increasing $n$, we may obtain less cases of collapsed favour graphs, and better relative entropies for the resulting rank distributions. Similarly, a finer-grained ranking from $k$-optimality is possible if we also take into account the *proportion* of $z$-objective subsets in which a point is efficient [10]. In both cases, the good performance (in at least some cases) of favour and $k$-optimality suggests that future study along these lines may be fruitful.

## 6. ACKNOWLEDGMENTS
Our thanks to the anonymous reviewers for their comments..

## 7. REFERENCES

[1] Bai, Z., Devroye, H., Hwang, H. and Tsai, T. Maxima in hypercubes. *Random Structures & Alg's* **27**:290—309, 2005.

[2] Bentley, J., Kung, H., Schkolnick, M. and Thompson, C. On the average number of maxima in a set of vectors and applications. *Journal of the ACM*, **25**(4):536—543, 1978.

[3] Bentley, P.J. and Wakefield, J.P. Finding acceptable solutions in the Pareto-optimal range using multiobjective genetic algorithms. In (Chawdry et al, eds.) *Soft Computing in Eng'g Design and Manufacturing*, Springer Verlag, 1997.

[4] Bonferroni, C.E. Il calcolo della assicurazioni su gruppi di teste. In *Studi in onore del Professore Salvatore Ortu Carbone*. Rome, Italy, pp. 13—60, 1935.

[5] Brockhoff, D. and Zitzler, E. Are all objectives necessary? On dimensionaility reduction in evolutionary multiobjective optimization. In *PPSN IX*, Springer LNCS, 533—542, 2006

[6] Coello, C.A.C. Handling preferences in evolutionary multiobjective optimization: a survey. In *Proc. of the 2000 Congress on Evo.. Computation*, vol 1, pp. 30—37, 2000.

[7] Corne, D.W., Knowles, J.D. and Oates, M.J. The Pareto-Envelope based selection algorithm for multiobjective optimization, in (Schoenauer et al, eds) *PPSN VI*, Springer LNCS pp. 869—878, 2000.

[8] Cvetkovic, D. and Parmee, I.C. Preferences and their application in evolutionary multiobjective optimization. *IEEE Trans. on Evol. Computation*, **6**(1):42—57, 2002.

[9] Deb, K. and Saxena, D.K. On finding Pareto-optimal solutions through dimensionality reduction for certain large-dimensional multi-objective optimization problems. *KanGAL Report No. 2005011*, IIT, Kanpur, 2005.

[10] di Pierro, F. Many-objective evolutionary algorithms and applications to water resources engineering. PhD thesis, University of Exeter, UK, August 2006.

[11] di Pierro, F., Djordjevic, S., Khu, S.-T, Savic, D. and Walters, G.A. Automatic calibration of urban drainage model using a novel multi-objective GA. In Krebs & Fuchs (eds.) *Urban Drainage Modelling'04*, pp. 41—52, 2004.

[12] Drechsler, D., Drechsler, R., Becker, B. Multi-objective optimisation based on relation *favour*. *In Proc. 1st EMO,* pp. 154—166, Springer Verlag, 2001.

[13] Farina, M. and Amato, P. A fuzzy definition of "optimality" for many-criteria optimization problems, *IEEE Trans SMC Part A*, **34**(3):315—326.

[14] Fonseca, C.M. and Fleming, P.J. Genetic algorithms for multiobjective optimization: formulation, discussion and generalization, In *Proc. 5th ICGA*, Morgan Kaufmann, pp. 416—423, 1993.

[15] Fonseca, C.M. and Fleming, P.J. Multiobjective optimization and multiple constraint handling with evolutionary algorithms. *IEEE Trans. SMC Part A*, **28**(1):26—37, 1998.

[16] Goldberg, D.E. *Genetic Algorithms in Search, Optimisation and Machine Learning*, Addison Wesley, 1989.

[17] Hanne, T. On the convergence of multiobjective evolutionary algorithms, *EJOR* **117**(3): 553—564, 1999.

[18] Hwang, C.L. and Yoon, K. *Multiple Attribute Decision Making: Methods and Applications, A State-of-the-Art Survey.* Springer, (1981)

[19] Knowles, J.D. and Corne, D.W. Approximating the nondominated front using the Pareto Archived Evolution Strategy. *Evolutionary Computation*, **8**(2):149—172, 2000.

[20] Knowles, J.D. and Corne, D.W. On metrics for comparing non-dominated sets. In *Proceedings of the 2002 Congress on Evolutionary Computation*, pp. 711—716, 2002.

[21] Maneeratana, K., Boonlong, K. and Chaiyaratana, N. Compressed-objective genetic algorithm, In *PPSN IX*, Springer LNCS, pp. 473—482, 2006.

[22] Purshouse, R. and Fleming, P.J. An adaptive divide-and-conquer strategy for evolutionary many-objective optimization. In *Proc. of 2nd Int'l Conf. on Evol. Multi-Criterion Optimization*, Srpringer, pp. 133—147, 2003.

[23] Schaffer, J.D. Multiple objective optimization with vector-evaluated genetic algorithms. In *Genetic algorithms and their applications: Proc. first Int'l Conf.*, pp. 93—100, 1985.

[24] Srinivas, N. and Deb, K. Multiobjective optimization using nondominated sorting genetic algorithm. *Evolutionary Computation*, **2**(3):221—248, 1994.

[25] Teytaud, O. How entropy theorems can show that offline approximating high-dim Pareto fronts is too hard. *Presented at PPSN BTP Workshop, PPSN 2006*, http://www.lri.fr/~teytaud/pareto2.pdf

[26] Van Veldhuizen, D.V. and Lamont, G.B. On measuring multiobjective evolutionary algorithm performance. In *Proc. CEC 2000*, vol 1, pp. 204—211, 2000.

[27] Zitzler, E. and Thiele, L. Multiobjective evolutionary algorithms: a comparative case study and the strength pareto approach. *IEEE Trans. Evol. Comp.*, **3**(4):257—271, 1999.

[28] Zitzler, E., Thiele, L., Laumanns, M., Fonseca, C.M., da Fonseca, V.G. Performance assessment of multiobjective optimizers: an analysis and review. *IEEE Transactions on Evolutuionary Computation*, **7**(2):117—132, 2003.